\pdfoutput=1

\documentclass[11pt]{article}

\usepackage[]{acl}

\usepackage{times}
\usepackage{latexsym}
\usepackage{amsmath, nccmath}
\usepackage{amsthm}
\usepackage{booktabs}
\usepackage{multirow}
\usepackage{graphicx}
\usepackage{wrapfig}
\usepackage{enumitem}
\usepackage{url}
\usepackage{cleveref}
\usepackage{wrapfig,lipsum,booktabs}
\usepackage[labelfont=bf]{caption}
\usepackage[ruled, vlined, linesnumbered]{algorithm2e}
\SetKwComment{Comment}{$\triangleright$\ }{}
\SetKwProg{Init}{Initialize}{}{}
\newcommand{\mc}[1]{\mathcal{#1}}
\usepackage{colortbl}

\definecolor{LightGreen}{HTML}{00994D}
\definecolor{LightRed}{HTML}{FF0000}

\crefformat{section}{\S#2#1#3}
\crefformat{subsection}{\S#2#1#3}
\crefformat{subsubsection}{\S#2#1#3}

\usepackage[T1]{fontenc}

\usepackage[utf8]{inputenc}

\usepackage{microtype}
\usepackage[normalem]{ulem} 
\usepackage[colorinlistoftodos]{todonotes}

\newcommand{\method}[1]{\textsc{Murmur}}

\usepackage{ulem}

\title{\method{}: Modular Multi-Step Reasoning for Semi-Structured Data-to-Text Generation}

\author{Swarnadeep Saha$^1$\thanks{\hspace{5pt}Work done during an internship at Meta AI.} \quad Xinyan Velocity Yu$^2$ \quad Mohit Bansal$^1$ \\ 
\quad \textbf{Ramakanth Pasunuru}$^2$ \quad \textbf{Asli Celikyilmaz}$^2$ \\
 $^1$UNC Chapel Hill \quad $^2$Meta AI \\
 \texttt{\{swarna, mbansal\}@cs.unc.edu} \\
 \texttt{\{velocityyu, rpasunuru, aslic\}@meta.com}
 }

\begin{document}
\maketitle
\begin{abstract}
Prompting large language models has enabled significant recent progress in multi-step reasoning over text. However, when applied to text generation from semi-structured data (e.g., graphs or tables), these methods typically suffer from low semantic coverage, hallucination, and logical inconsistency. We propose \method{}, a neuro-symbolic modular approach to \textit{text generation from semi-structured data with multi-step reasoning}. \method{} is a best-first search method that generates reasoning paths using: (1) neural and symbolic modules with specific linguistic and logical skills, (2) a grammar whose production rules define valid compositions of modules, and (3) value functions that assess the quality of each reasoning step. We conduct experiments on two diverse data-to-text generation tasks like WebNLG and LogicNLG. These tasks differ in their data representations (graphs and tables) and span multiple linguistic and logical skills. \method{} obtains significant improvements over recent few-shot baselines like direct prompting and chain-of-thought prompting, while also achieving comparable performance to fine-tuned GPT-2 on out-of-domain data. Moreover, human evaluation shows that \method{} generates highly faithful and correct reasoning paths that lead to 26\% more logically consistent summaries on LogicNLG, compared to direct prompting.

\end{abstract}

\begin{figure}[t]
    \centering
    \includegraphics[width=\columnwidth]{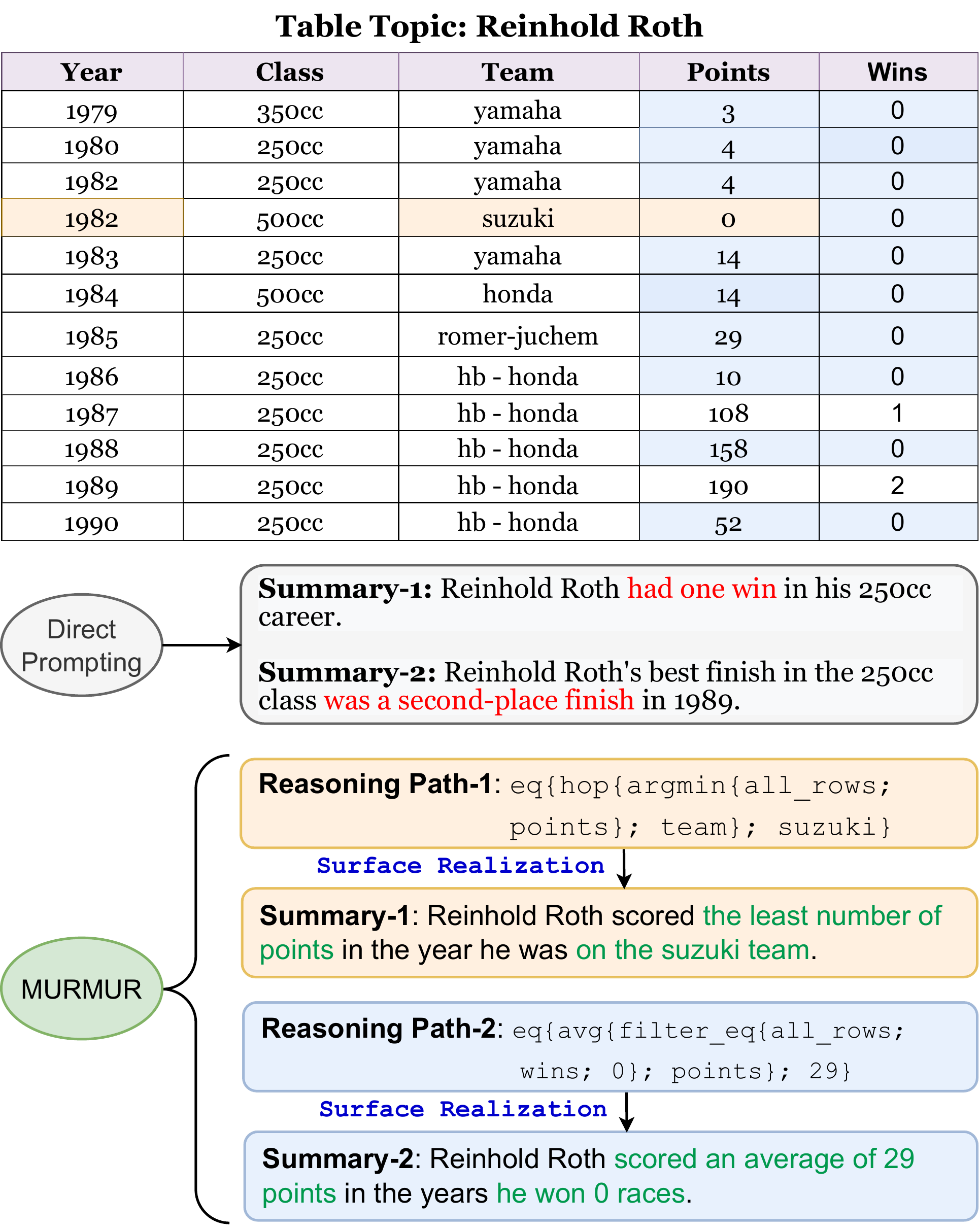}
    \caption{
    Sample table from LogicNLG dataset and two logical summaries generated by \method{} and Direct Prompting baseline. Direct Prompting summaries include logical inconsistencies and hallucinations (marked in \textcolor{LightRed}{red}) while \method{} generates reasoning paths (composed of several modules) and converts them to logically consistent summaries (marked in \textcolor{LightGreen}{green}). Each color code in the table cells highlights part of the table relevant to a \method{} summary.
    }
    \vspace{-5pt}
    \label{fig:top_fig}
\end{figure}

\section{Introduction}
Data-to-text generation~\cite{mckeown1992text, reiter1997building, wen2015semantically, duvsek2015training, mei2016talk, novikova2017e2e, gatt2018survey} is the task of generating fluent, faithful, and consistent summaries of semi-structured data. Recent works have introduced different data-to-text generation tasks wherein the data is represented in diverse structures, like meaning representations~\cite{novikova2017e2e}, graphs~\cite{gardent2017webnlg}, or tables~\cite{lebret2016neural, parikh2020totto, chen2020logical}. Text generation from such data is challenging because it extends surface realization of the input content and requires various reasoning and compositionality skills, such as filtering a table based on a certain criterion, retrieving the maximum value from a table column, etc.

Recent works fine-tune pre-trained language models~\cite{radford2019language, raffel2020exploring} as the de-facto standard for building supervised data-to-text generation systems~\cite{kale2020text, agarwal2021knowledge}.
However, this requires a large amount of domain-specific parallel data, which is expensive to obtain, and training models on such data also affects out-of-domain generalization~\cite{laha2020scalable, duvsek2020evaluating}.

Motivated by the recent success of few-shot prompting in multi-step reasoning over text~\cite{wei2022chain, nye2021show, wang-etal-2022-robust, dohan2022language}, we pose data-to-text generation as \textit{multi-step reasoning over data}.\footnote{By data, we mean semi-structured data such as graphs or tables. Henceforth, we will refer to it as just `data'.} However, reasoning over data brings its own set of challenges, especially in the context of text generation: (1) \textbf{Generation Quality}: Firstly, directly prompting large language models (LLMs) can cause models to suffer from low semantic coverage, hallucinations, and logically inconsistent generations when reasoning over complex forms of data, like tables (see red marked phrases for the Direct Prompting summaries in Fig.~\ref{fig:top_fig}). Other prompting methods like Chain-of-Thought (CoT) encourage LLMs to also generate intermediate reasoning steps~\cite{wei2022chain}. However, it compromises the \textit{transparency}, \textit{faithfulness},\footnote{Faithful reasoning refers to an underlying causal structure in the reasoning process. This is different from a generated text's faithfulness to an input context which henceforth, will be referred to as hallucinations.} and \textit{correctness} of the reasoning process due to the lack of explicit conditioning between the reasoning steps~\cite{creswell2022faithful}. (2) \textbf{Transformation-invariance:} Unlike text, which is a sequence of tokens, data is typically represented as a \textit{set} of elements (e.g., a graph is a set of edges while a table is a set of rows, as shown in Fig.~\ref{fig:top_fig}). Hence, a model that reasons over data must be \textit{transformation-invariant}~\cite{wang-etal-2022-robust}. For instance, the summary generated from a table should be invariant to randomly shuffling the rows of the table. Thus, prompting methods that linearize the data in an arbitrary order, can be prone to some variance (see Table~\ref{tab:webnlg_test} and \ref{tab:logicnlg_test}). 

We propose \method{}, a \textbf{m}od\textbf{u}la\textbf{r} \textbf{mu}lti-step \textbf{r}easoning approach to text generation from data (\cref{sec:murmur}). It is a best-first search algorithm (\cref{sec:algo}) that generates reasoning paths (see examples in Fig~\ref{fig:top_fig}) with three features: (1) \textbf{Modularity} (\cref{sec: modules}): \method{} defines a set of neural and symbolic modules with diverse input/output data types that constitute multiple steps in a reasoning path. Neural modules perform linguistic skills that LLMs are good at (e.g., the \textit{Surface Realization} module in Fig.~\ref{fig:top_fig} converts a reasoning path to a natural language summary) and symbolic modules perform logical skills that they mostly struggle with \cite{wang2022behavior, gao2022pal} (e.g., the \textit{argmin} module in Fig.~\ref{fig:top_fig} finds the row with the minimum points); (2) \textbf{Grammar} (\cref{sec:grammar}): \method{} introduces a grammar whose production rules specify valid compositions of modules. For instance, in the second path of Fig.~\ref{fig:top_fig}, \method{} first generates the module \textit{filter\_eq} followed by the \textit{avg} module, because the former outputs a \textit{table} data type which is also the input data type to the latter; (3) \textbf{Value functions} (\cref{sec:value_func}): To evaluate the quality of each plausible reasoning step and choose the best modules at each step, \method{} defines value functions that score, rank, and select the best steps. For example, in the second path of Fig.~\ref{fig:top_fig}, an \textit{avg} module is perhaps more salient than a \textit{max} or \textit{min} module (which only finds the maximum or minimum points).

Our \textbf{findings} are: \method{} can perform multi-step generative reasoning on simple to complex semi-structured data-to-text generation tasks including WebNLG~\cite{gardent2017webnlg}, a graph-to-text task (\cref{sec:g2t_exp}) and LogicNLG~\cite{chen2020logical}, a table-to-text task (\cref{sec:t2t_exp}). We compare \method{} with state-of-the-art supervised (end-to-end and pipeline) and few-shot prompting methods. On WebNLG, \method{} obtains significant improvements in semantic coverage and hallucinations of generated summaries over other few-shot baselines like direct prompting and CoT prompting. Additionally, \method{} demonstrates good out-of-domain generalizability by obtaining comparable performance to fine-tuned LMs like GPT-2. On LogicNLG, human evaluations demonstrate that \method{} significantly improves the logical consistency of summaries over direct prompting (by up to 26\%), showcasing the strength of a neuro-symbolic approach for data-to-text generation.

\begin{figure*}[t]
    \centering
    \includegraphics[width=\textwidth]{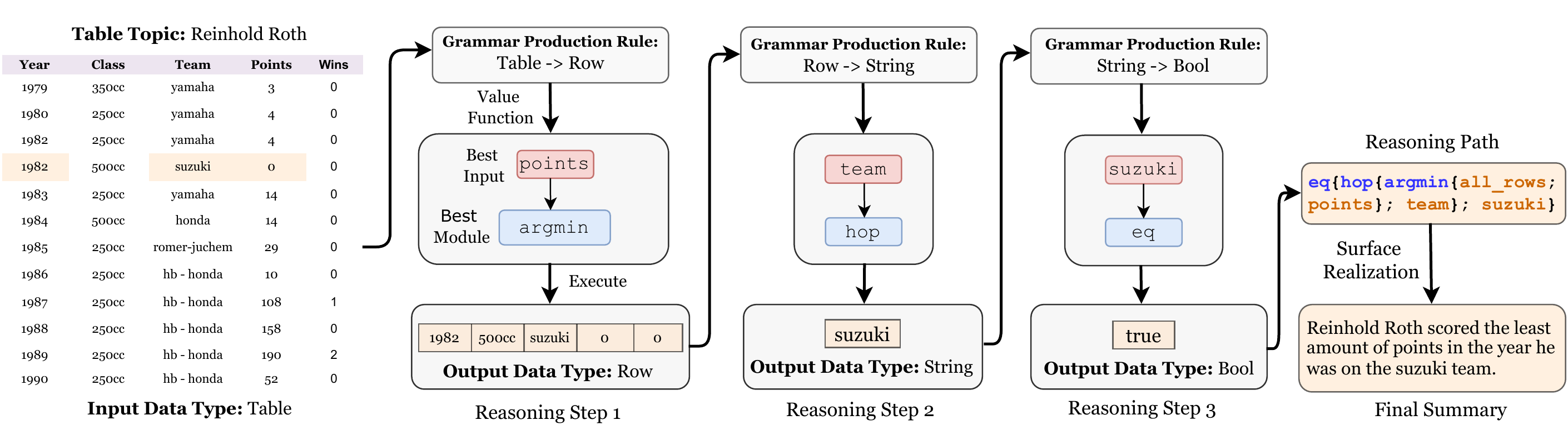}
    \caption{Illustration of \method{} generating a reasoning path and then converting it into a logically consistent summary, supported by the input table. The reasoning path consists of three reasoning steps. At each step, \method{} chooses a set of plausible modules (according to a grammar) and then selects the best module, with the best input according to a value function. The output generated at each step serves as the input to the next step.
    }
    \vspace{-5pt}
    \label{fig:model_fig}
\end{figure*}
\section{Definitions: Reasoning Step and Path}
We begin by formally defining a \textit{Reasoning Step} and a \textit{Reasoning Path}. A Reasoning Step is a triple ($\mc{M}, \mc{X}, y)$ where a module $\mc{M}$ performs a certain skill by conditioning on an input $\mc{X}$ to generate an output $y$. For example, in Fig.~\ref{fig:model_fig}, the module \textit{argmin} takes a table and a column (\textit{points}) as input and outputs the row with the minimum points. A \textit{Reasoning Path} is defined as a sequence of such reasoning steps $\{(\mc{M}_i, \mc{X}_i, y_i)\}_{i=1}^r$. Fig.~\ref{fig:model_fig} shows an example of a reasoning path, represented as a nested structure. It consists of three reasoning steps for three modules (\textit{argmin}, \textit{hop}, and \textit{eq}). The \textit{argmin} module outputs the row in the table with minimum points, which is the input to the next module \textit{hop} that selects a column from that row. \method{} generates textual summaries from semi-structured data by constructing such reasoning paths that are then converted to the final outputs through a \textit{Surface Realization} module, as shown in Fig.~\ref{fig:model_fig}.\footnote{For better illustration, we show Surface Realization outside of the reasoning path but ideally, it can be considered as another (final) step in the reasoning path.}

\section{\method{} Approach}
\label{sec:murmur}
\method{} consists of four components: (1) a set of modules, (2) a grammar, (3) value function(s), and (4) a search algorithm that brings all the previous three components together. The search algorithm constructs reasoning paths by first identifying plausible modules at each reasoning step according to the grammar and then determines the best modules and their corresponding inputs with the help of value functions. Fig.~\ref{fig:model_fig} shows a working example of \method{}, in which it starts with an input table, searches for a reasoning path (composed of three steps) and finally converts it into a summary. The specifics of \method{}'s components vary based on the task at hand. As case studies, we consider two data-to-text generation tasks. The first one is WebNLG~\cite{gardent2017webnlg}, a graph-to-text generation task. The goal of the task is to generate a natural language description of the input graph (represented as a set of triples). The second task is LogicNLG~\cite{chen2020logical}, a complex table-to-text generation task where the goal is to generate logically consistent summaries from salient parts of the table. We now describe each component.

\subsection{\method{} Modules}
\label{sec: modules}
\method{} defines a set of modules $\{\mc{M}_i\}_{i=1}^m$ that perform specialized reasoning skills for the corresponding task. Formally, each module $\mc{M}_i$ is defined as a multi-variate function $\mc{M}_i : \mc{X}$ $\rightarrow y $ that maps an \textit{n}-tuple input $\mc{X}$ = ($x_1$,$\cdots$, $x_n$) to an output $y$. Each input variable $x_i$ and output $y$ can have their own expected data types $d_i$ and $d_y$ respectively. These data types could be user-defined\footnote{The modules are analogous to function definitions in a programming language with expected IO types. Similarly, user-defined data types can be thought of as class definitions. For instance, a data type \textit{Triple} can be implemented as a class consisting of a subject, a relation, and an object (all with data type \textit{String}).} like \textit{Table}, \textit{Triple}, etc or primitive ones like \textit{String}, \textit{Number}, \textit{Bool}, etc. For example, in Fig.~\ref{fig:model_fig}, the module $\mc{M}_{argmin} : (t, c) \rightarrow r$ takes a table $t$ (with data type \textit{Table)} and a column $c$ (with data type \textit{String}) as input and outputs a row $r$ (with data type \textit{row}) with the minimum value in column $c$. 

The modules are implemented as few-shot neural models or symbolic functions. We choose neural modules for linguistic skills that LLMs typically excel at and symbolic modules for logical operations that LLMs mostly struggle with~\cite{wang2022behavior, gao2022pal}. Reasoning over semi-structured data allows us to define these symbolic modules that can be easily implemented with \textsc{Python} functions. Below we provide examples of neural and symbolic modules for the two tasks.

\paragraph{Neural Linguistic Modules.} In any modular data-to-text generation approach, one of the modules is responsible for the transition from structured data to unstructured text. We call it Surface Realization. In particular, for WebNLG, we define it as $\mc{M}_{\mathit{sr}} : t \rightarrow s$ that converts a triple $t$ (with data type \textit{Triple}) into a short sentence $s$ (with data type \textit{String}). For LogicNLG, we define it as $\mc{M}_{\mathit{sr}} : (t, p) \rightarrow s$ that takes a table $t$ (with data type \textit{Table}) and a reasoning path $p$ as input and converts it into a summary $s$ (with data type \textit{String}). As we show later, in WebNLG, \textit{Surface Realization}s are the first reasoning steps, in LogicNLG, it is the last step and more generally, can be any one of the steps for an arbitrary data-to-text reasoning task. The transition between the reasoning steps with different data types is explicitly learned with the help of a grammar (\cref{sec:grammar}).

For WebNLG, we also define a \textit{Text Fusion} module $\mc{M}_{\mathit{tf}} : (s_1, s_2) \rightarrow s$ that combines two pieces of text $s_1$ and $s_2$ (with data types \textit{String}) into a coherent text $s$. \textit{Text Fusion} performs fusion in an iterative fashion by combining a pair of intermediate generations at each step. This enables controllability of the generated content yielding more coherent generations~\citep{tan-etal-2021-progressive}.

\begin{table}[h]

 \begin{center}
	   \scalebox{0.74}
    {
     \begin{tabular}{>{\raggedright}p{2cm}p{7.4cm}}
    \toprule
         \textbf{Module} & \textbf{Description}\\
    \midrule
        {Filtering} &
        $\mc{M}_{filter} : (t, cr) \rightarrow t'$: takes a table $t$ and a filtering criterion $cr$ as input and outputs a table $t'$ with rows where the criterion $cr$ is satisfied. \\ 
    \midrule
        {Aggregation} & Performs different aggregation operations on a table. For example, $\mc{M}_{max} : (t, c) \rightarrow n$ is a \textit{max} module that takes a table $t$ and a column $c$ as input and outputs the maximum number $n$ in column $c$.\\
    \midrule
        {Boolean}  & $\mc{M}_{bool} : (t, cr) \rightarrow b$: takes a table $t$ and a criterion $cr$ as input and outputs a boolean $b$ based on whether the criterion is satisfied.  \\
    \midrule
        {Hop} & $\mc{M}_{hop} : (r, c) \rightarrow e$: takes a row $r$ and a column $c$ as input and outputs the element $e$ in column $c$ for the row.\\
    \bottomrule
    \end{tabular}
    }
    \vspace{-5pt}
    \caption{\method{} Symbolic Modules to perform logical operations over tables in Table-to-Text generation. }
	\label{tab:symmodules}
    \vspace{-10pt}
	\end{center}
\end{table}
\paragraph{Symbolic Logical Modules.} For LogicNLG, drawing motivation from prior work~\cite{chen2020logic2text}, we define different categories of symbolic modules that perform logical operations over tables (see Table~\ref{tab:symmodules} and refer to Table~\ref{tab:appendix_logicnlg_modules} for the detailed list). WebNLG requires summarizing an input graph and hence, does not involve any logical modules.

\begin{table}[]\centering
\small
\begin{tabular}{l}
\toprule
\multicolumn{1}{c}{Graph-to-Text (WebNLG)} \\ \midrule
         Triple $\rightarrow$ String \\
         (String, String) $\rightarrow$ String \\ \midrule
         \multicolumn{1}{c}{Table-to-Text (LogicNLG)} \\ \midrule
Table $\rightarrow$ Table | Row | Number | Boolean \\
Row $\rightarrow$ String | Number \\
String | Number $\rightarrow$ Boolean    \\ 
(Table, Path) $\rightarrow$ String \\ \bottomrule
\end{tabular}
\caption{Grammars for WebNLG and LogicNLG defining production rules between different data types.}
\label{tab:grammar}
\end{table}

\subsection{Grammar over Modules}
\label{sec:grammar}
The role of the grammar is to determine a set of plausible modules in a reasoning step and how they should be composed. The production rules of the grammar capture possible transitions from an input data type to an output data type(s) (see Fig.~\ref{fig:model_fig} and Table~\ref{tab:grammar}). Each production rule thus defines multiple permissible modules. For example, the production rule `Table $\rightarrow$ Number' (meaning that a number can be generated from a table) is valid for both \textit{max} and \textit{min} modules. When \method{} searches for reasoning paths, the grammar reduces the search space (over all possible modules) by only selecting the ones that can be composed at each reasoning step. We provide examples below of how such grammars are constructed.

\paragraph{Grammar for Graph-to-Text Generation.} Table~\ref{tab:grammar} contains the grammar for Graph-to-Text generation. It consists of two production rules. The surface realization module takes an input of type \textit{Triple} and outputs a \textit{String}. The text fusion module takes an input of a pair of \textit{Strings} and outputs a \textit{String}. Past pipeline approaches for graph-to-text generation~\cite{jiannan2022ASDOT} also perform surface realization followed by fusion, as explained through the grammar.

\paragraph{Grammar for Table-to-Text Generation.}Generating logical summaries from a table is a more challenging task because it involves a large number of modules. Based on the types of modules introduced previously, we define a grammar, as shown in Table~\ref{tab:grammar}. The production rules encode the possible transitions between different data types. For example, the first rule encodes the knowledge that given an input of type \textit{Table}, one can output a \textit{Table}, a \textit{Row} of the table, a \textit{Number}, or a \textit{Boolean}. The final rule encodes the surface realization module that converts a reasoning path to a summary.

\subsection{Value Functions}
\label{sec:value_func}
While the grammar helps reduce the search space by defining permissible compositions of modules, each reasoning step can still have multiple plausible modules and each module can also have multiple plausible inputs to choose from. Thus, \method{} introduces value functions (see Fig.~\ref{fig:model_fig}) that assess the quality of each plausible reasoning step (module and input to it) by scoring, ranking, and selecting the best reasoning step(s).

\paragraph{Value Function for Graph-to-Text Generation.} In a Graph-to-Text generation task, each intermediate reasoning step $r$ generates a summary $y_r$ for a subset of edges (triples) $G_r$ from the input graph (see Fig.~\ref{fig:appendix_webnlg_generative_process} for an illustration). \method{} introduces value functions that assess the following two aspects of the generated summary $y_r$.

\noindent\textbf{Fluency} is measured by log-likelihood of the generated text similar to BARTScore~\cite{bartscore}: 
\begin{equation}
    S_f (y_r) = \exp\{\frac{1}{l}\sum_{i=1}^{l}\log p_\theta (y_r^i|y_r^{<i})\} \nonumber
\label{eqn:value_fluency}
\end{equation}
\noindent\textbf{Semantic Consistency} measures the average logical entailment probability $\mathit{P}_e(\cdot)$ between the generated text $y_r$ and input triples $G_r$\footnote{We concatenate the surface realizations of the triples to construct the sequence for the NLI model.} and vice-versa:
\begin{equation}
    S_{sc} (G_r, y_r) = 0.5 \times (P_e (G_r, y_r) + P_e (y_r, G_r)) \nonumber
\label{eqn:value_consistency}
\end{equation}
We use an NLI model to infer entailment probabilities. The both-way entailment scores capture equivalence between the triples and the generation, ensuring that the latter not only covers all the triples but also does not hallucinate any new information. 
Overall score $S(\cdot)$ is an ensemble of the two scores:
\begin{equation}
S(G_r, y_r) = \alpha S_f (y_r) + (1-\alpha) S_{sc} (G_r, y_r) \nonumber
\label{eqn:value_overall}
\end{equation}
\paragraph{Value Function for Table-to-Text Generation.} For these tasks, our value function can be used to choose the best module(s) at each reasoning step, as well as the best input(s) for the corresponding module(s).\footnote{In Graph-to-Text generation, we do not need to choose the best module because at each step there is only one plausible module. Hence, the value function is only aimed at determining the best inputs to the \textit{Text Fusion} module.} For instance, if a reasoning step generates a number from a table (according to the grammar), the value function should determine the best module(s) between \textit{max}, \textit{min}, etc, as well as which column should the \textit{max} or \textit{min} module be operating on. Taking inspiration from past work on verifying intermediate reasoning traces over text~\cite{creswell2022faithful, yang2022generating}, we train a value function $S: (T, P_r) \rightarrow p$ that can judge the correctness of an intermediate or partial reasoning path $P_r$ for an input table $T$. In particular, we train a binary classification model on samples with correct and incorrect partial reasoning paths. We call this value function a \textit{saliency metric} because it selects the best reasoning steps that reason over salient parts of the table. We discuss the model and training data for our saliency metric in $\S$~\ref{sec:exp_setup_logicnlg}.

\subsection{\method{} Search Algorithm}
\label{sec:algo}
We now describe how all the three components discussed above (modules, grammar, and value functions) come together in generating reasoning paths for \method{} (see Fig.~\ref{fig:model_fig}). It is a best-first search algorithm that operates as follows: The algorithm takes as input a set of $m$ modules $\{\mc{M}_i\}_{i=1}^{m}$, a grammar $\mc{G}$, a value function $\mc{V}$, and number of reasoning paths or summaries to generate $p$. Additionally, it considers a hyperparameter, the beam size $b$ of the search ($b \geq p$). The search begins by initializing an empty priority queue that maintains the beam (best $b$ partial reasoning paths to be explored anytime during the search). Next, at each step, \method{} (1) pops an element from the queue, (2) identifies the data type of the element (e.g., \textit{Table}), (3) looks up the grammar to find all possible transitions from that data type (e.g., \textit{Row}, \textit{Number}), (4) selects all modules for each such transition (e.g., \textit{argmax} and \textit{argmin} for `Table $\rightarrow$ Row', \textit{max} and \textit{min} for `Table $\rightarrow$ Number'), and (5) constructs all plausible reasoning steps consisting of modules and their corresponding inputs (e.g., all numerical columns for \textit{argmax}). It then scores all these reasoning steps using the value function, ranks them, and only keeps the top-$b$ paths in the queue. For WebNLG, the search terminates when all triples have been iterated. For LogicNLG, a reasoning path is complete when the module at the current step outputs a boolean variable that evaluates to true (e.g., the \textit{eq} module). Upon termination of the search, we return the top-$p$ paths and the corresponding summaries.

\section{Experimental Setup}
\label{experimentsetup}

\subsection{Graph-to-Text Generation}

WebNLG~\cite{gardent2017webnlg} contains RDF triples from DBPedia~\cite{auer2007dbpedia}.\footnote{We experiment with the 2017 version of the dataset available at \url{https://webnlg-challenge.loria.fr/challenge_2017/}.}  The test split consists of two parts -- the first half contains DB categories seen in training data, and the second half contains 5 unseen categories, which are used to evaluate model generalization. 

\paragraph{Modules.} We implement both modules, \textit{Surface Realization} and \textit{Text Fusion} as few-shot neural models by prompting a large language model, OPT-175B~\cite{zhang2022opt} with skill-specific prompts (see Appendix~\ref{sec:prompts}). We perform greedy decoding to generate text from each module.

\begin{table*}[]
\small
\centering
\begin{tabular}{clcccccc} 
\toprule
 & & \multicolumn{3}{c}{BLEU} & \multicolumn{3}{c}{METEOR} \\ 
 & & Seen      & Unseen      & All      & Seen       & Unseen       & All      \\ \midrule
 \parbox[t]{2mm}{\multirow{4}{*}{\rotatebox[origin=c]{90}{supervised}}}& MELBOURNE$^\dagger$ & 54.5 & 33.2 & 45.1 & 41.0 & 33.0 & 37.0 \\
& GPT-2-large$^\dagger$ & 65.3 & 43.1 & 55.5 & 46.0 & 38.0 & 42.0     \\
& T5-large$^\dagger$ & 64.9 & 54.0 & 59.9 & 46.0 & 43.0 & 44.0 \\
& Neural Pipeline$^\ddagger$ & - & - & 43.3 & - & - & 39.3    \\ \midrule
\parbox[t]{2mm}{\multirow{4}{*}{\rotatebox[origin=c]{90}{few-shot}}}& Direct Prompting (k=1)$^\star$ & 33.1$\pm$0.3  & 34.2$\pm$0.1 & 33.6$\pm$0.1 & 30.4$\pm$0.1 & 31.2$\pm$0.1 & 30.8$\pm$0.1 \\
& Direct Prompting (k=5)$^\star$ & 39.9$\pm$0.3 & 38.9$\pm$0.3 & 39.5$\pm$0.1 & 34.3$\pm$0.1 & 34.3$\pm$0.3 & 34.4$\pm$0.1  \\
& CoT Prompting (k=1)$^\star$ & 22.2$\pm$0.2 & 14.9$\pm$0.2 & 18.0$\pm$0.1 & 22.3$\pm$0.1 & 22.9$\pm$0.2 & 22.6$\pm$0.1\\
& \method{} (k=1)$^\star$ & \textbf{41.4$\pm$0.0} & \textbf{41.1$\pm$0.0} & \textbf{41.3$\pm$0.0} & \textbf{37.1$\pm$0.0} & \textbf{37.1$\pm$0.0} & \textbf{37.1$\pm$0.0} \\
\bottomrule
\end{tabular}
\caption{Comparison of supervised (end-to-end and pipeline) and few-shot approaches on the WebNLG Seen and Unseen splits of the test set. $\dagger$ = Supervised with 7k in-domain samples. $\ddagger$ = Supervised with a synthetic corpus of 934k samples. $\star$ = Few-shot with $k$ demonstrations. We report mean and variance for all few-shot methods with three random orderings of the input triples.}
\label{tab:webnlg_test}
\end{table*}

\paragraph{Value Function.} As part of the value function described in \cref{sec:value_func}, we compute fluency score using the log probabilities estimated by the OPT-175B model. The entailment probability for the semantic scorer is based on a state-of-the-art DeBERTa-base model~\cite{he2020deberta} trained on a collection of eight NLI datasets.\footnote{The model is publicly available at \url{https://huggingface.co/MoritzLaurer/DeBERTa-v3-base-mnli-fever-docnli-ling-2c}.} We choose the mixing ratio $\alpha$ between the two scorers to be $0.05$. At each reasoning step, \method{} scores and ranks the intermediate generations in the queue using the value function. Subsequently, it only explores the highest scoring intermediate generation in the next step of the search and prunes the rest.\footnote{This is equivalent to performing greedy search. We experimented with larger beams for a more exhaustive search without observing any noticeable improvement in performance.}

\subsection{Table-to-Text Generation}
\label{sec:exp_setup_logicnlg}

\paragraph{Modules.} We implement all logical modules, as described in \cref{sec: modules}, with \textsc{Python} functions. We prompt OPT-175B for the \textit{Surface Realization} module that converts a reasoning path into a natural language summary (see Appendix~\ref{sec:prompts} for the prompt).

\paragraph{Value Function.} Our saliency metric is a binary classifier. Specifically, we train a BERT-base model that takes a table (linearized into a sequence of rows) and a partial reasoning path as input and classifies it as correct or incorrect. During inference, we consider the correct class probability as the saliency score. We obtain training data for our model from the Logic2Text dataset~\cite{chen2020logic2text} that annotates open-domain tables with gold reasoning paths. Given a gold reasoning path, we create \textit{correct} partial paths by breaking it at each intermediate step/module and \textit{incorrect} paths by performing two types of perturbations on every correct partial path: (1) replacing the module at the current step with another module of same data type (e.g., replacing module \textit{max} with module \textit{min}); (2) replacing the inputs to the module with other plausible inputs (e.g., replacing \textit{max} over column $c_1$ with \textit{max} over column $c_2$). See Fig.~\ref{fig:saliency_data} and \cref{sec:app_saliency_metric} for an illustration of the training data creation process. We choose 221 (table, reasoning path) pairs from the Logic2Text dataset and convert them into 1500 correct and incorrect training samples consisting of (table, partial reasoning path) pairs. While choosing the samples, we ensure that the corresponding tables have no intersection with those in the test and validation sets of LogicNLG. We choose the beam size of the search to be 20 (see further analysis of varying beam sizes in \cref{sec:app_logicnlg_beam}).

\section{Experiments on Graph-to-Text Generation}
\label{sec:g2t_exp}
We discuss our experimental findings on the WebNLG dataset.
\subsection{Comparison of \method{} with supervised and few-shot methods}

\paragraph{Baselines.} As supervised methods (trained on 7k examples), we experiment with \textbf{MELBOURNE}, a non-pretrained encoder-decoder model~\cite{gardent2017webnlg} and two fine-tuned pretrained language models, \textbf{GPT-2-large}~\cite{radford2019language} and \textbf{T5-large}~\cite{raffel2020exploring}. We also compare with a state-of-the-art modular pipeline approach, \textbf{Neural Pipeline}~\cite{kasner2022neural} that first converts triples to sentences using hand-designed templates and subsequently orders and fuses the sentences by fine-tuning on a large synthetic corpus of 934k samples. Lastly, for direct comparisons, we consider two few-shot baselines, \textbf{Direct Prompting (DP)} that directly prompts the OPT-175B model to generate a summary from the graph, and \textbf{Chain-of-Thought Prompting (CoT)}~\cite{wei2022chain} that prompts the model to generate the summary step-by-step (see \cref{sec:prompts} for example prompts). We choose the demonstrations randomly from the training data and kept them consistent across all few-shot methods.

\paragraph{Metrics.} Following prior work, we perform automatic evaluation using BLEU~\cite{papineni2002bleu} and METEOR~\cite{banerjee2005meteor}. On the test set, we report results both on the seen and unseen splits.

\paragraph{Results.} For all few-shot methods, we report mean and variance of three randomly chosen orderings of the input triples. Table~\ref{tab:webnlg_test} shows the results, demonstrating the following:

\begin{itemize}[nosep, wide=0pt, leftmargin=*, after=\strut]
\item \method{} significantly outperforms DP and CoT by up to 8 points in BLEU and METEOR ($p < 0.001$), when using a single demonstration (k=1).\footnote{A single demonstration for DP can get decomposed into multiple in-context examples in \method{}. However, like CoT, the decompositions i.e., the intermediate reasoning steps are still part of the same demonstration.} \method{} even outperforms DP with five demonstrations (k=5). We observe that CoT significantly underperforms DP because the intermediate reasoning steps have limited coherence between them and prompting a LLM by concatenating all steps does not work well for text generation. Through its modular treatment of the task, \method{} breaks the single example down into individual reasoning steps, thereby allowing the model to first convert each triple into a sentence and then combine two pieces of text in the successive reasoning steps. In the process, \method{} generates outputs with more coverage of triples and lesser hallucinations, as reflected in the improved scores and further demonstrated in \cref{sec:webnlg_human} through human evaluation. 

\item \method{} outperforms a non-pretrained supervised baseline MELBOURNE and obtains comparable performance to fine-tuned LMs like GPT-2 on the unseen test split, thereby showcasing the out-of-domain generalizability of our method. All supervised methods exhibit a drop in performance from the seen to the unseen split because of in-domain training, while \method{} shows robust generalization behavior. 

\item \method{} is transformation-invariant (i.e., its generations are invariant to the order of triples) because it treats the graph as a \textit{set} (not \textit{sequence}) of triples while all other prompting methods operate by linearizing the graph into a randomly chosen order of the triples. Lack of transformation-invariance yields up to 0.3 points variance in BLEU scores for DP and CoT baselines.
\end{itemize}

\begin{table}[]
\small
\centering
\begin{tabular}{lcc | c} 
\toprule
 & DP & \method{} & \% Improve \\ \midrule
Omissions$\downarrow$ & 1.64$\pm$0.06 & \bf 0.73$\pm$0.01 & +24\% \\
Hallucinations$\downarrow$ & 0.77$\pm$0.03 & \bf 0.43$\pm$0.03 & +9\% \\
Redundancies$\downarrow$ & 0.00$\pm$0.00 & 0.00$\pm$0.00 & - \\
Disfluencies$\downarrow$ & \bf 0.14$\pm$0.05 & 0.30$\pm$0.04 & -4\% \\
\bottomrule
\end{tabular}
\caption{Average count of omissions, hallucinations, redundancies, and disfluencies in the final outputs generated by DP and \method{} on WebNLG. We report mean and variance of the counts between the two annotators. Percentage improvement is computed with respect to the average number of triples (3.8) in the input samples.}
\label{tab:webnlg_human_first}
\end{table}

\subsection{Human Evaluation of Final Generations and Intermediate Reasoning Steps}
\label{sec:webnlg_human}

We conduct two steps of human evaluations. First, following prior work~\cite{kasner2022neural, jiannan2022ASDOT}, we compare the final summaries generated by DP (our best baseline) and \method{}. Second, we also evaluate the faithfulness and correctness of each individual reasoning step generated by \method{}.

\paragraph{Comparison of DP and \method{} final summaries.} Two NLP experts take part in the study with 50 randomly chosen test samples. We ask the annotators to count the number of omissions, hallucinations, redundancies, and disfluencies (grammatical issues) in the generated outputs and report mean and variance of the averaged counts between the two annotators.\footnote{All metrics including hallucinations and disfluencies are counted at the level of triples or facts. This allows us to compare the raw counts of both methods with respect to the average number of triples in the input samples.} The test samples contain an average of 3.8 triples. Our results in Table~\ref{tab:webnlg_human_first} demonstrate that \method{} significantly reduces the average omissions of triples from 1.64 to 0.73, leading to an absolute improvement of 24\%. Similarly, there is a significant 9\% reduction of average hallucinations with \method{}. We attribute these improvements to \method{}'s step-wise generative process and selecting the best intermediate generations through our value function. Compared to DP, we observe a slight drop in fluency in \method{}'s generations because of the step-wise fusion operations while DP generates the summaries in one go. In \cref{sec:appendix_examples}, we show examples of summaries generated by DP and \method{}.

\begin{table}[t]
\small
\centering
\begin{tabular}{cccc} 
\toprule
Module & Grammatical & Faithful & Correct \\ \midrule
Surface Realization & 1.00 & 1.00 & 0.82 \\
Text Fusion & 0.90 & 0.72 & 0.64 \\
\bottomrule
\end{tabular}
\caption{Fraction of grammatical, module faithful, and correct intermediate reasoning steps generated by the two modules in \method{} for WebNLG.}
\label{tab:webnlg_faithfulness}
\end{table}
\vspace{-.5em}
\paragraph{Faithfulness and Correctness of Intermediate Reasoning Steps of \method{}.} Correct final generations do not guarantee \textit{module faithfulness} or correctness of the intermediate reasoning steps~\cite{subramanian2020obtaining}. Hence, in addition to evaluating final generated summaries, we also evaluate the quality of the individual reasoning steps of \method{}. For every reasoning step of a data point, we provide the annotators with (1) step level generations of \method{}, and (2) the previous steps that the current step is conditioned on (i.e., the corresponding triple for a \textit{surface realization} module and the two texts for a \textit{fusion} module). We conduct this study on six randomly chosen test examples, spanning 50 reasoning steps (28 Surface Realization and 22 Text Fusion samples). We judge the generation at each reasoning step for its grammaticality, module faithfulness (i.e., if the module is doing what it is supposed to do, e.g., whether the fusion operation actually fuses two pieces of information), and correctness (e.g., whether the fusion is correct). From Table~\ref{tab:webnlg_faithfulness}, we observe that both modules generate outputs that are almost always grammatical. Module faithfulness is also significantly high -- all surface realization modules resolve the triples and text fusion also involves some form of fusion a high 72\% of the times. 64\% of the fusion generations are also fully correct.

\subsection{Effect of Number of Demonstrations}
\label{effectofdenomonstrations}
\begin{figure}[t]
    \centering
    \includegraphics[width=\columnwidth]{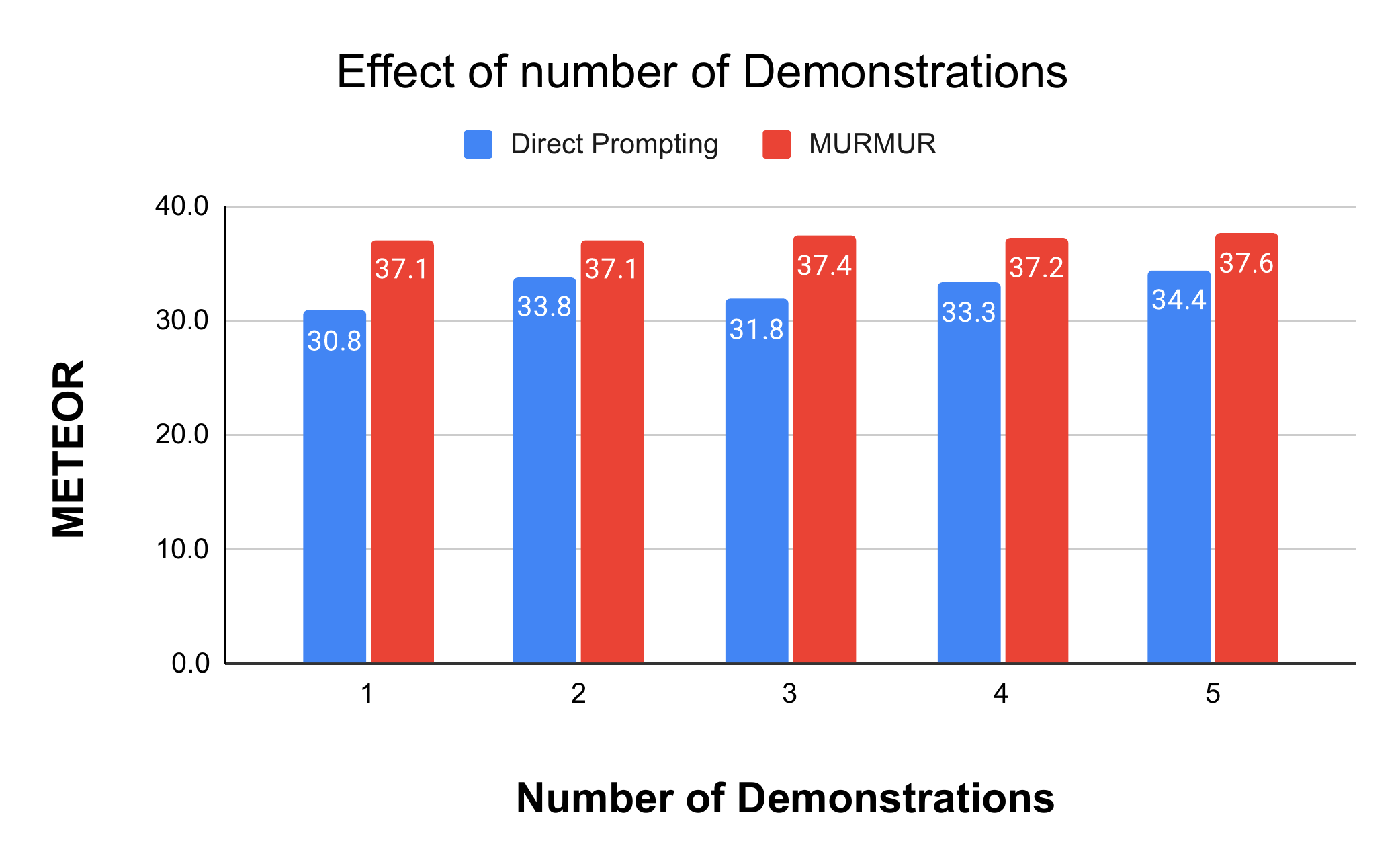}
    \caption{Comparison of METEOR scores for Direct Prompting (DP) versus \method{} with varying number of demonstrations on WebNLG test set. DP shows improved performance with more demonstrations while \method{}'s gains are marginal.}
    \label{fig:plot_webnlg}
\end{figure}
In Fig.~\ref{fig:plot_webnlg}, we compare the METEOR scores of DP and \method{} by varying the number of demonstrations. DP shows improved performance with more demonstrations, while \method{}'s improvements are marginal. In the process of providing more demonstrative examples, DP implicitly learns the underlying step-wise reasoning process, while such phenomenon is explicitly captured through one demonstration in \method{}. We also observe that \method{} is robust to variations in the choice of demonstrations (see details in \cref{sec:appendix_webnlg}).

\section{Experiments on Table-to-Text Generation}
\label{sec:t2t_exp}
Below are our findings on the LogicNLG dataset.

\subsection{Comparison of \method{} with supervised and few-shot methods}

\paragraph{Baselines.} We compare \method{} to several state-of-the-art supervised methods from prior work which are trained on 37k (table, summary) pairs using supervised training. These include a non-pretrained transformer model, \textbf{Field-Infusing + Trans}, three pre-trained LMs based on BERT and GPT-2, \textbf{BERT-TabGen}, \textbf{GPT-TabGen}, \textbf{GPT-Coarse-to-Fine}~\cite{chen2020logical} and a de-confounded variational encoder-decoder model, \textbf{DCVED}~\cite{chen2021confounded}. Similar to WebNLG, we also compare with \textbf{Direct Prompting} and \textbf{CoT Prompting}. Additionally, we evaluate the effect of our search algorithm and value function (saliency metric) by conducting two ablations. In \textbf{BART + SR}, instead of generating the reasoning paths through search, we fine-tune a BART model that generates reasoning paths in one go. As training data, we leverage the (table, gold reasoning path) pairs that are used for training the metric in \method{}. The surface realization (SR) step of converting the path to summary is left unchanged. In the second ablation, we remove the saliency metric from \method{} by selecting the module (and its inputs) at each step randomly (but according to the grammar). All few-shot methods use 1 randomly chosen demonstration from the training data.

\paragraph{Metrics.} Following prior work~\cite{chen2020logical}, we compare all methods with BLEU scores. \citet{chen2020logical} also propose some logical fidelity metrics like using an NLI model to evaluate logical consistency between the summary and the table. However, we found such learned metrics do not correlate well with humans and instead, we conduct more reliable human evaluations for evaluating logical correctness  of \method{} (\cref{sec:logicnlg_human}).

\begin{table}[]
\centering
\small
\begin{tabular}{lc}
\toprule
             & BLEU-1 / BLEU-2 / BLEU-3 \\ \midrule
Field-Infusing$^\dagger$ & 43.7 / 20.9 / 8.4 \\
BERT-TabGen$^\dagger$ & 49.1 / 27.7 / 13.5 \\
GPT-TabGen$^\dagger$ & 49.6 / 28.2 / 14.2 \\
GPT-Coarse-to-Fine$^\dagger$ & 49.0 / 28.3 / 14.6 \\
DCVED$^\dagger$ & 49.5 / 28.6 / 15.3 \\ \midrule
Direct Prompting$^\star$ & 37.2$\pm$0.4 / 18.8$\pm$0.2 / 8.6$\pm$0.2 \\
CoT Prompting$^\star$ & 35.6$\pm$0.2 / 18.6$\pm$0.1 / 8.8$\pm$0.0 \\
BART + SR$^\ddagger$ & 39.2$\pm$0.2 / 20.6$\pm$0.2 / 9.5$\pm$0.0\\
\method{}$^\ddagger$ & \textbf{39.8$\pm$0.0} / \textbf{22.2$\pm$0.0} / \textbf{11.2$\pm$0.0} \\
- saliency$^\star$ & 39.6$\pm$0.0 / 21.9$\pm$0.0 / 10.6$\pm$0.0 \\
\bottomrule 
\end{tabular}
\caption{Comparison of supervised non-pretrained, pretrained, and few-shot approaches on the LogicNLG test set. $\dagger$ = Supervised with 37k in-domain samples. $\star$ = Few-shot with 1 demonstration. $\ddagger$ = Few-shot with 1 demonstration and 221 gold (table, path) pairs. We report mean and variance for all few-shot methods with three random orderings of the input table rows.}
\label{tab:logicnlg_test}
\end{table}

\begin{table*}[ht]
\small
\centering
\begin{tabular}{ccccc | c} 
\toprule
& Correct & Partial & Incorrect & Ungrammatical &  Is Logical?  \\ \midrule
Direct Prompting & 28.7$\pm$3.7 & 20.0$\pm$2.5 & 38.8$\pm$8.7 & 12.5$\pm$2.5 & 62.0$\pm$0.5 \\
\method{} & \textbf{55.0$\pm$2.5} &	1.2$\pm$1.2 & 38.8$\pm$3.7 & 5.0$\pm$2.5 & \bf 95.4$\pm$0.2 \\
\bottomrule
\end{tabular}
\caption{Human evaluation for LogicNLG showing the percentages of fully correct, partially correct, incorrect, and ungrammatical generations from Direct Prompting and \method{}. `Is Logical' denotes the percentage of correct generations that involve some underlying logical computations (as opposed to just surface-level). We report mean and variance of the annotations between two experts.}
\label{tab:logicnlg_human}
\end{table*}
\paragraph{Results.} Table~\ref{tab:logicnlg_test} shows the results on the test set of LogicNLG. We draw the following conclusions.

\begin{itemize}[nosep, wide=0pt, leftmargin=*, after=\strut]
\item \method{} significantly improves ($p < 0.001$) upon direct prompting and CoT prompting by up to 2.4 points in BLEU-3. We attribute this to two factors: (1) leveraging symbolic modules for logical skills that ensure their correctness, (2) delegating the task of converting a path to a natural language summary to OPT-175B, a linguistic task that LLMs are mostly good at. Unlike DP, \method{} conditions on the searched paths to control the generations which yields logically consistent generations, as further demonstrated through human evaluation in \cref{sec:logicnlg_human}. CoT, while generating an intermediate reasoning path, does not use executable modules and hence cannot guarantee valid compositionality, logical correctness or faithfulness in the reasoning steps. 

\item \method{} improves upon the Field-Infusing model, a non-pretrained transformer model by 2.8 BLEU-3 points. It also closes the gap to state-of-the-art supervised models to 4 BLEU-3 points.

\item Unlike DP and CoT prompting, \method{} is unaffected by content-invariant transformations of a table like randomly shuffling the rows of the table because all symbolic modules treat the table as a set of rows. We conduct other variance analyses like choice and number of demonstrations on all few-shot methods in \cref{sec:app_logicnlg_var_demo} and \cref{sec:app_logicnlg_num_demo} respectively.

\item We observe 0.6 points improvement in BLEU-3 score with the saliency metric, indicating that it helps in choosing more salient reasoning paths. Even without this metric, all reasoning paths generated by \method{} are still valid because of its grammar component. In \cref{sec:app_saliency_metric}, we analyze the effect of varying the amount of supervision for the saliency metric on downstream LogicNLG performance. 

\item Finally, we observe a significant drop in performance when our search algorithm is replaced with a fine-tuned BART model (BART + SR). This model is the fine-tuned equivalent of CoT and like CoT, is prone to generating reasoning paths that are neither logically correct or faithful.
\end{itemize}

\subsection{Human Evaluation of Logical Correctness}
\label{sec:logicnlg_human}

Next, we conduct human evaluation to assess the logical correctness of the generations from Direct Prompting and \method{}. Two NLP experts annotate 40 randomly chosen generations from eight different tables. In particular, they take part in two studies. First, they classify each generation into whether it is ungrammatical, grammatical but incorrect, grammatical but partially correct, or grammatical and also fully correct. Next, for each correct generation (from both models), they annotate whether the generation involves any underlying logical operation (like counting, summation, etc) or are mere surface realizations of the table content. We report the results in Table~\ref{tab:logicnlg_human}. We observe that \method{} not only generates 26\% more correct outputs, but about 95\% of those generations also involve some logical operations. Generating reasoning paths through symbolic modules ensures that almost all generations are logical derivations from the table, an ability that is significantly harder to achieve through direct prompting. Refer to~\cref{sec:appendix_examples} for some logical summaries generated by \method{}.

\section{Related Work}

\subsection{Multi-step Reasoning over Text} 

Recent developments in large language models~\cite{brown2020language, zhang2022opt, thoppilan2022lamda, chowdhery2022palm} have enabled significant progress in few-shot methods for logical reasoning tasks~\cite{wei2022chain, creswell2022selection, nye2021show, wang2022self, zelikman2022star, zhou2022least, dasgupta2022language, kojima2022large, dohan2022language}. Representative methods like chain-of-thought prompting encourage language models to output intermediate reasoning steps before generating the final answer~\cite{wei2022chain}. However, the reasoning steps are all generated in one go from a single model, potentially leading to unfaithful reasoning due to the lack of explicit conditioning between the steps~\cite{creswell2022faithful}. \method{} overcomes this issue by developing granular modules that are capable of performing specialized skills by \textit{explicitly} conditioning on the outputs from previous reasoning steps. Conceptually, \method{} bears similarity with the Selection-Inference architecture~\cite{creswell2022selection, creswell2022faithful} in which at each reasoning step, a selection module selects relevant facts to reason over and an inference module generates an inference from the selected facts. However, their focus is on question answering and reasoning over textual context~\cite{saha2020prover, saha2021multiprover, tafjord2021proofwriter, dalvi2021explaining, bostrom2022natural}. A few concurrent works have also proposed neuro-symbolic approaches for reasoning over text via generating programs and further executing them through symbolic solvers~\cite{gao2022pal, wang2022behavior, chen2022program, cheng2022binding}. Different from these works, we tackle a more challenging setup of multi-step reasoning in the context of \textit{controlled generation} and over \textit{semi-structured data}.

\subsection{Modular Reasoning over Text}

\method{} follows a body of work on Neural Module Networks~\cite{andreas2016neural, jiang2019self, gupta2019neural, subramanian2020obtaining, saha2021weakly} that learn and execute compositional programs over modules. While the modules in a Neural Module Network typically output attention maps, prior works have also used text-in text-out modules whose input and output data types are \textit{strings}~\cite{khot2021text, khot2022decom, saha2022summarization}. \method{}'s modules are a generalization of text-in text-out modules since they can capture operations with diverse signatures including complex data types (like \textit{tables}) and \textit{strings}, among others. The transition from data to text is also clearly represented through the compositions of our modules, unlike attention maps-based modules whose interpretability has often been debated~\cite{serrano2019attention}.

\subsection{Data-to-Text Generation} 
Existing methods for data-to-text generation can be broadly grouped into three categories -- (1) supervised methods that finetune seq2seq pre-trained language models on large amounts of parallel data~\cite{kale2020text, chen2020kgpt, ribeiro2021investigating, ke2021jointgt, jiannan2022ASDOT}, (2) pipeline approaches that leverage different modules~\cite{reiter1997building, reiter2007architecture, laha2020scalable, kasner2022neural}, and (3) few-shot methods that assume access to a large corpus of unlabeled examples for data augmentation or retrieving similar examples~\cite{puduppully2019data, zhao2020bridging, trisedya2020sentence, su2021plan}. Unlike prior modular approaches, \method{} uses few-shot neural or symbolic modules that do not require any manual intervention. Unlike past few-shot methods, \method{} works well with as few as one demonstration, without requiring access to any unlabeled corpus.

\section{Discussion and Conclusion}

We presented \method{}, a neuro-symbolic modular reasoning approach for data-to-text generation. Through extensive experiments on two tasks, WebNLG and LogicNLG, we demonstrated that \method{} significantly outperforms few-shot baselines such as Direct Prompting and Chain-of-Thought Prompting, achieves comparable performance to fine-tuned LMs like GPT-2, and generates significantly more logical summaries. 

\method{} shows the benefits of building interpretable modular text generation systems by breaking a task down into sub-problems and then solving them through separate modules, without requiring module-specific supervision. It utilizes the power of large language models in solving linguistic sub-tasks through in-context learning, while delegating the logical sub-tasks to symbolic modules. \method{} generalizes the concept of modules by treating them as functions and defining their behaviours through expected input/output data types. This, in turn, facilitates the introduction of a grammar for explaining module compositions (analogous to function compositions). Future work could explore extending \method{} for text generation tasks that involve reasoning over multiple modalities, such as text, semi-structured data, and images.

\bibliography{custom}

\appendix

\begin{table*}[]
\small
\begin{tabular}{p{0.15\linewidth} | p{0.2\linewidth} | p{0.17\linewidth} | p{0.4\linewidth}}
\toprule
\textbf{Module Name} & \textbf{Input Data Type} & \textbf{Output Data Type} & \textbf{Description} \\ \midrule \midrule
    filter\_eq \hspace{100pt} filter\_not\_eq & table, string, string & table & Returns a table with the rows where entry in the input column (second argument) is equal or not equal to the input value (third argument).\\ \midrule
    
    filter\_greater \hspace{100pt} filter\_greater\_eq \hspace{100pt} filter\_lesser \hspace{100pt} filter\_lesser\_eq & table, string, number & table & Returns a table with the rows where a numerical column (second argument) is greater than or less than (or equal to) the input number (third argument).\\ \midrule
    
    filter\_all & table, string & table & Returns the whole table. \\ \midrule \midrule
    
    arg\_max \hspace{100pt} arg\_min & table, string & row & Returns the row with the minimum or maximum value for the input column (second argument). \\ \midrule
    
    max \hspace{100pt} min \hspace{100pt} avg \hspace{100pt} sum & table, string & number & Returns the maximum, minimum, average or sum of numbers in the input column (second argument). \\ \midrule
    
    count & table & number & Returns the number of rows in the table. \\ \midrule \midrule
    
    all\_eq \hspace{100pt} all\_not\_eq & table, string, string & bool & Returns whether all entries in the input column are equal (or not equal to) the input value. \\\midrule
    
    all\_greater \hspace{100pt} all\_less \hspace{100pt} all\_greater\_eq \hspace{100pt} all\_less\_eq  & table, string, number & bool & Returns whether all entries in the input column are greater than or less than (or equal to) the input number. \\ \midrule
    
    most\_eq \hspace{100pt} most\_not\_eq & table, string, string & bool & Returns whether most entries in the input column are equal (or not equal to) the input value. \\\midrule
    
    most\_greater \hspace{100pt} most\_less \hspace{100pt} most\_greater\_eq \hspace{100pt} most\_less\_eq & table, string, number & bool & Returns whether most entries in the input column are greater than or less than (or equal to) the corresponding number. \\ \midrule
    
    only & table & bool & Returns whether the table has exactly one row. \\ \midrule \midrule
    
    hop & row, string & string & Returns the entry corresponding to the input column in the row. \\ \midrule \midrule
    
    eq & string, string & bool & Returns whether the two inputs are equal or not. \\
    \bottomrule
\end{tabular}
\caption{List of modules for LogicNLG with their corresponding input / output data types and descriptions.}
\label{tab:appendix_logicnlg_modules}
\end{table*}

\section{Modules for Table-to-Text Generation (Cont. from \cref{sec: modules})}

Table~\ref{tab:appendix_logicnlg_modules} shows the list of all modules for LogicNLG. Our choice of modules is motivated from prior work~\cite{chen2020logic2text} that defines similar modules for generating logical summaries from open-domain tables.
\section{Additional Experiments on WebNLG (Cont. from \cref{sec:g2t_exp})}
\label{sec:appendix_webnlg}

\subsection{Effect of Variations of Demonstrations}

In Table~\ref{tab:webnlg_dev}, we compare the performance of few-shot baselines on the validation set of WebNLG and analyze the effect of different choices of random demonstrations on in-content learning. Using three different random seeds, we show that all methods are fairly robust to randomness in demonstrations.
\begin{table}[h]
\small
\centering
\begin{tabular}{lcc} 
\toprule
 & BLEU & METEOR \\ \midrule
Direct Prompting (k=1) & 31.1$\pm$0.5 & 29.8$\pm$0.1\\
Direct Prompting (k=5) & 38.3$\pm$0.4 & 33.6$\pm$0.1\\
\method{} (k=1) & \textbf{40.1$\pm$0.3} & \textbf{37.1$\pm$0.5} \\
\bottomrule
\end{tabular}
 \vspace{-5pt}
\caption{Comparison of different few-shot methods on the WebNLG validation set. We report mean and variance of BLEU and METEOR scores with three different random seeds for choosing demonstrations from the training set.}
\label{tab:webnlg_dev}
\end{table}

\section{Additional Experiments on LogicNLG (Cont. from \cref{sec:t2t_exp})}

\subsection{Effect of Variations in Demonstrations}
\label{sec:app_logicnlg_var_demo}

In Table~\ref{tab:logicnlgdev}, we study the effect of randomness in the choice of a single demonstration for LogicNLG. We report mean and variance of BLEU scores for each method with a randomly chosen demonstration from the training examples. Similar to WebNLG, all methods are fairly robust to the choice of demonstrations and exhibit comparable variance in performance.
\begin{table}[h]
\centering
\small
\begin{tabular}{lccc}
\toprule
             & BLEU-1 / BLEU-2 / BLEU-3 \\ \midrule
Direct Prompting (k=1) & 37.0$\pm$0.2 / 18.9$\pm$0.1 / 8.5$\pm$0.1	 \\
COT Prompting (k=1) & 36.5$\pm$0.1 / 18.9$\pm$0.1 / 8.7$\pm$0.3 \\
\method{} (k=1) & \textbf{40.5$\pm$0.1} / \textbf{22.2$\pm$0.0} / \textbf{10.8$\pm$0.1} \\
\bottomrule 
\end{tabular}
 \vspace{-5pt}
\caption{Comparison of different few-shot methods on the LogicNLG validation set. We report mean and variance of BLEU scores with two random seeds for choosing one demonstration from the training set.}
\label{tab:logicnlgdev}
\end{table}

\subsection{Effect of Number of Demonstrations}
\label{sec:app_logicnlg_num_demo}

\begin{figure}[t]
    \centering
    \includegraphics[width=\columnwidth]{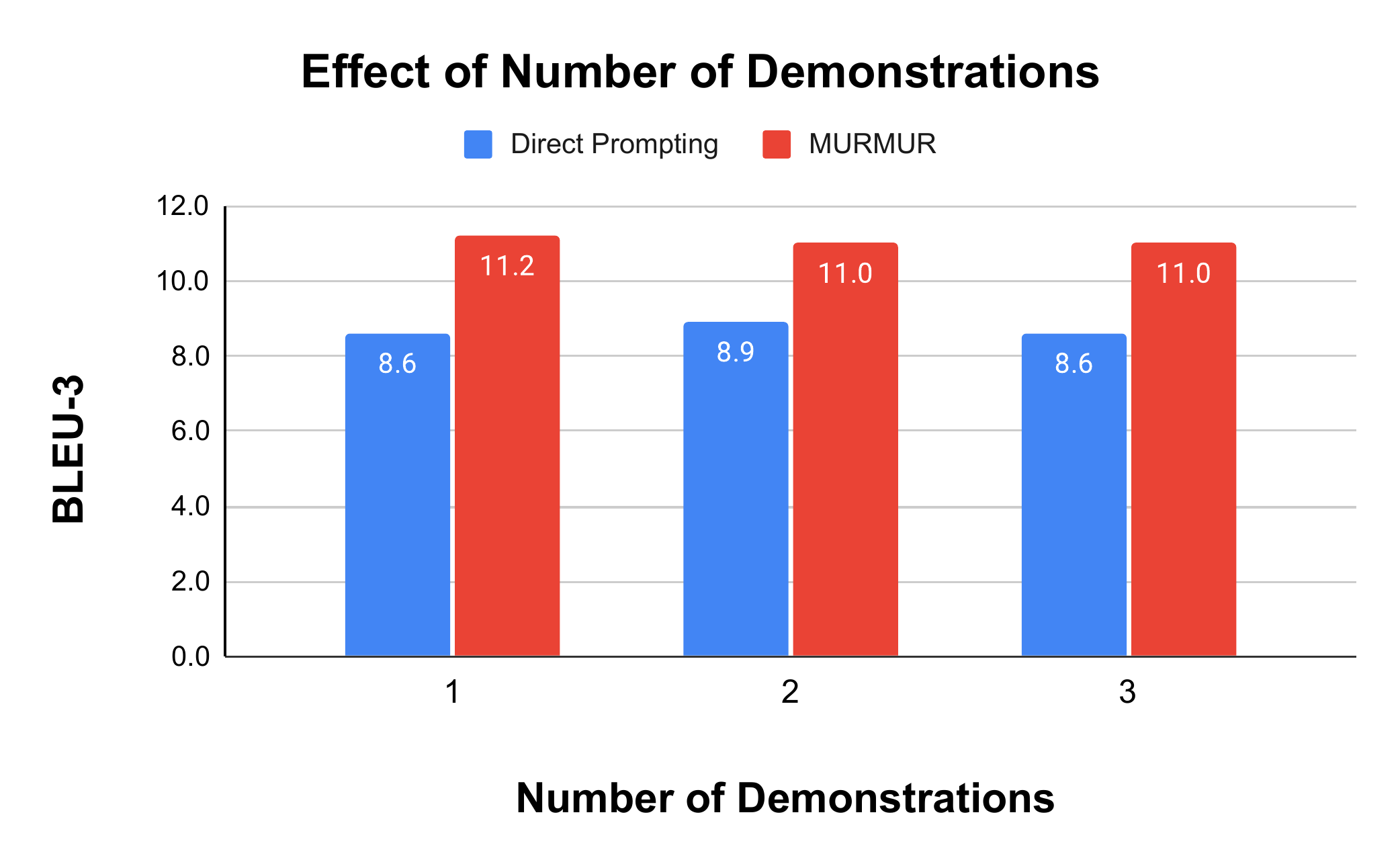}
     \vspace{-5pt}
    \caption{Comparison of BLEU-3 scores for Direct Prompting versus \method{} with varying number of demonstrations on LogicNLG validation set. For both methods, results do not improve further with more demonstrations.}
    \label{fig:plot_logicnlg}
\end{figure}
In Fig.~\ref{fig:plot_logicnlg}, we compare BLEU-3 scores of DP and \method{} by varying the number of demonstrations from 1 to 3. Unlike WebNLG, we do not observe any noticeable improvements in in-context learning capabilities with more demonstrations, possibly because of the inherent difficulty of generating logical summaries from tables.

\subsection{Effect of Different Beam Sizes in Best-first Search of \method{}}
\label{sec:app_logicnlg_beam}
At each step of the search, \method{} keeps track of the highest scoring reasoning paths. Table~\ref{tab:logicnlg_dev_beam} compares the effect of the beam size for our search algorithm on the LogicNLG validation set. Perhaps unsurprisingly, maintaining a bigger beam i.e., conducting a more exhaustive search leads to some improvements in BLEU scores, however, the gain mostly saturates with beam sizes of around 50-100.
\begin{table}[t]
\centering
\small
\begin{tabular}{cccc}
\toprule
       Beam Size      & BLEU-1 / BLEU-2 / BLEU-3 \\ \midrule
10 & 39.7 / 21.2 / 10.3	 \\
20 & 40.2 / 21.8 / 10.7 \\
50 & 40.5 / 22.2 / 10.8 \\
100 & \bf 40.7 / \bf 22.5 / \bf 10.9 \\
\bottomrule 
\end{tabular}
\caption{Effect of beam size in \method{}'s search algorithm on BLEU scores of LogicNLG validation set.}
\label{tab:logicnlg_dev_beam}
\end{table}

\begin{figure*}[t]
    \centering
    \includegraphics[width=\textwidth]{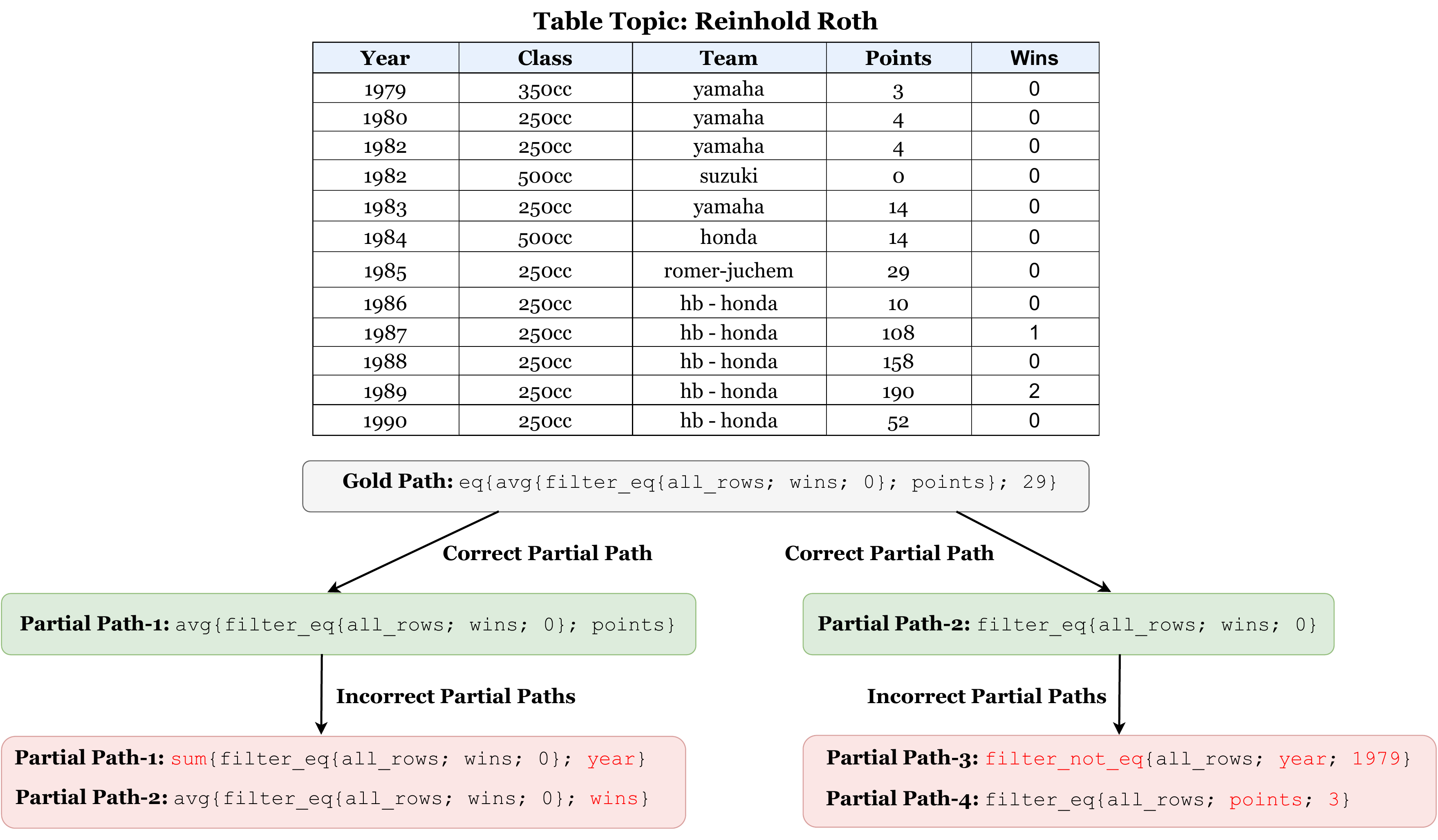}
    \caption{Training data creation process for the saliency metric. Given a gold path, we create correct (positive) partial paths by breaking the path at each step. From each correct partial path, we create incorrect partial paths by doing two kinds of perturbations, one at the module level and another at the inputs level (Cont. from \cref{sec:exp_setup_logicnlg}).
    }
    \vspace{-5pt}
    \label{fig:saliency_data}
\end{figure*}
\subsection{Further Analysis of Saliency Metric (Cont. from \cref{sec:exp_setup_logicnlg})}
\label{sec:app_saliency_metric}
\paragraph{Training Data Construction.} In Fig.~\ref{fig:saliency_data}, we show an illustrative example of the training data creation process for our saliency metric. In `Incorrect Partial Path-1', when we perturb the \textit{avg} module with the \textit{sum} module, we aim to teach the model that although both are valid reasoning steps, averaging over the column `points' is a more salient and informative reasoning step than summing over the column `year'. Similarly, in `Incorrect Partial Path-2', when we perturb the input to the module \textit{avg} by performing average over the column `wins', we want the model to learn the salient columns to reason over for a given module.

\paragraph{Effect of Varying Supervision on Metric Accuracy and Downstream Performance.} We conduct an in-depth analysis of the saliency metric used to score the reasoning steps in \method{}. As shown in Table~\ref{tab:logicnlg_saliency}, we vary the amount of supervision for training the saliency metric and study its effect on the validation set accuracy (in identifying whether a partial reasoning path is correct or not) and also on the downstream LogicNLG BLEU scores. Our key takeaway is that a small number of gold reasoning paths (about 200, spanning 100 tables) is enough to train a good saliency metric that not only achieves a high classification accuracy of 76\% but also leads to a BLEU-3 score of 10.8 on LogicNLG. Increasing the training data further to 7k gold paths (equivalently, 42k correct and incorrect partial paths) increases the classification accuracy to 82\% but does not impact LogicNLG performance much.
\begin{table}[h]
\centering
\small
\begin{tabular}{@{}ccccc@{}}
\toprule
          \# Gold  & \# Gold  & \# Samples  & Acc. & LogicNLG  \\
           Tables &  Paths & (Pos/Neg/ All) &  &  BLEU-3 \\ \midrule
100   & 221 &  769/729/1498 & 76.16 & 10.8  \\
200 & 443 & 1534/1457/2991 & 78.52  & 10.6        \\
500 & 1085 & 3773/3633/7406 & 80.32 & 10.9  \\
3000 & 7145 & 21.5k/20.5k/42.0k & 82.84 & 10.7 \\ \bottomrule 
\end{tabular}
\caption{Effect of varying amount of supervision for the saliency metric on the metric accuracy (Acc.) and on downstream LogicNLG BLEU scores. Metric accuracy is computed on 4.4k validation samples consisting of 2264 correct paths (positive samples) and 2179 incorrect paths (negative samples).}
\label{tab:logicnlg_saliency}
\end{table}

\section{Prompts (Cont. from \cref{experimentsetup})}
\label{sec:prompts}

\paragraph{WebNLG.} Table~\ref{tab:direct_prompt_webnlg} shows an example of direct prompting~\cite{zhang2022opt} for WebNLG. In Table~\ref{tab:sr_prompt_webnlg} and Table~\ref{tab:fusion_prompt_webnlg}, we show the prompts for the \textit{surface realization} and ~\textit{text fusion} modules in \method{}. Note that the single demonstration for direct prompting is decomposed into individual reasoning steps for the two modules in \method{}.

\begin{table}[h]
\small
    \centering
    \scalebox{0.9}{
    \begin{tabular}{p{\linewidth}}
         \toprule
         Let's convert triples to sentences \\
         \#\#\# \\
         Triples: \textcolor{blue}{A.S.\_Gubbio\_1910 | league | Serie\_D \# Italy | leader | Pietro\_Grasso \# Italy | capital | Rome \# A.S.\_Gubbio\_1910 | ground | Italy \# Serie\_D | champions | S.S.\_Robur\_Siena} \\
         Output: \textcolor{olive}{S.S. Robur Siena are champions of Serie D in which AS Gubbio 1910 also play. This latter club have their home ground in Italy where the capital city is Rome and the leader is Pietro Grasso.} \\
         \#\#\# \\
         Triples: \textcolor{blue}{\{triples\}} \\
         Output:\\
         \bottomrule
    \end{tabular}}
    \vspace{-5pt}
    \caption{Example of Direct Prompting for WebNLG.}
    \label{tab:direct_prompt_webnlg}
\end{table}

\begin{table}[h]
\small
    \centering
    \scalebox{0.9}{
    \begin{tabular}{p{\linewidth}}
         \toprule
         Let's convert triples to sentences step-by-step \\
         \#\#\# \\
         Triples: \textcolor{blue}{A.S.\_Gubbio\_1910 | league | Serie\_D \# Italy | leader | Pietro\_Grasso \# Italy | capital | Rome \# A.S.\_Gubbio\_1910 | ground | Italy \# Serie\_D | champions | S.S.\_Robur\_Siena} \\
         Output: \textcolor{olive}{AS Gubbio 1910 plays in Serie D. \# Pietro Grasso is the leader of Italy. \# ... \# S.S. Robur Siena are champions of Serie D in which AS Gubbio 1910 also play. This latter club have their home ground in Italy where the capital city is Rome and the leader is Pietro Grasso.} \\
         \#\#\# \\
         Triples: \textcolor{blue}{\{triples\}} \\
         Output:\\
         \bottomrule
    \end{tabular}}
    \vspace{-5pt}
    \caption{Example of Chain-of-Thought Prompting for WebNLG. The intermediate reasoning steps (truncated for clarity) are concatenated together and we consider the last step as the final summary.}
    \label{tab:cot_prompt_webnlg}
\end{table}

\begin{table}[h]
\small
    \centering\scalebox{0.9}{
    \begin{tabular}{p{\linewidth}}
         \toprule
         Let's convert a triple to a sentence \\
         \#\#\# \\
         Triple: \textcolor{blue}{A.S.\_Gubbio\_1910 | league | Serie\_D} \\
         Sentence: \textcolor{olive}{AS Gubbio 1910 plays in Serie D.} \\
         \#\#\# \\
         Triple: \textcolor{blue}{Italy | leader | Pietro\_Grasso} \\
         Sentence: \textcolor{olive}{Pietro Grasso is the leader of Italy.} \\
         \#\#\# \\
         Triple: \textcolor{blue}{Italy | capital | Rome} \\
         Sentence: \textcolor{olive}{Rome is the capital of Italy.} \\
         \#\#\# \\
         Triple: \textcolor{blue}{A.S.\_Gubbio\_1910 | ground | Italy} \\
         Sentence: \textcolor{olive}{Italy is the home ground of AS Gubbio 1910.} \\
         \#\#\# \\
         Triple: \textcolor{blue}{Serie\_D | champions | S.S.\_Robur\_Siena} \\
         Sentence: \textcolor{olive}{S.S. Robur Siena are champions of Serie D.} \\
         \#\#\# \\
         Triple: \textcolor{blue}{\{triple\}} \\
         Sentence:\\
         \bottomrule
    \end{tabular}}\vspace{-5pt}
    \caption{Example of the Surface Realization prompt for \method{} in WebNLG.}
    \label{tab:sr_prompt_webnlg}
\end{table}

\begin{table}[h]
\small
    \centering
    \scalebox{0.9}{
    \begin{tabular}{p{\linewidth}}
         \toprule
         Let's combine two sentences \\
         \#\#\# \\
         First Sentence: \textcolor{blue}{S.S. Robur Siena are champions of Serie D.} \\
         Second Sentence: \textcolor{cyan}{AS Gubbio 1910 plays in Serie D.} \\
         Combined Sentence: \textcolor{olive}{S.S. Robur Siena are champions of Serie D in which AS Gubbio 1910 also play.} \\
         \#\#\# \\
         First Sentence: \textcolor{blue}{Rome is the capital of Italy.} \\
         Second Sentence: \textcolor{cyan}{Pietro Grasso is the leader of Italy.} \\
         Combined Sentence: \textcolor{olive}{Rome is the capital of Italy where Pietro Grasso is the leader.} \\
         \#\#\# \\
         First Sentence: \textcolor{blue}{S.S. Robur Siena are champions of Serie D in which AS Gubbio 1910 also play.} \\
         Second Sentence: \textcolor{cyan}{Italy is the home ground of AS Gubbio 1910.} \\
         Combined Sentence: \textcolor{olive}{S.S. Robur Siena are champions of Serie D in which AS Gubbio 1910 also play. This latter club have their home ground in Italy.} \\
         \#\#\# \\
         First Sentence: \textcolor{blue}{S.S. Robur Siena are champions of Serie D in which AS Gubbio 1910 also play. This latter club have their home ground in Italy.} \\
         Second Sentence: \textcolor{cyan}{Rome is the capital of Italy where Pietro Grasso is the leader.} \\
         Combined Sentence: \textcolor{olive}{S.S. Robur Siena are champions of Serie D in which AS Gubbio 1910 also play. This latter club have their home ground in Italy where the capital city is Rome and the leader is Pietro Grasso.} \\
         \#\#\# \\
         First Sentence: \textcolor{blue}{\{sent1\}} \\
         Second Sentence: \textcolor{cyan}{\{sent2\}} \\
         Combined Sentence:\\
         \bottomrule
    \end{tabular}}
    \vspace{-5pt}
    \caption{Example of the Text Fusion prompt for \method{} in WebNLG.}
    \label{tab:fusion_prompt_webnlg}
\end{table}

\paragraph{LogicNLG.} Table~\ref{tab:direct_prompt_logicnlg} shows an example of direct prompting for LogicNLG. Table~\ref{tab:sr_prompt_logicnlg} shows an example prompt for the \textit{surface realization} module in LogicNLG. We only provide the table topic, table header, and the reasoning path in the prompt. We do not add the table content to the prompt because all the information needed by the model to generate the summary is typically present in the reasoning path. Any other contextual information about the table can also be inferred from the table header and topic. We observe that adding the table content makes the model more prone to hallucinations because it may not limit its generation to the information provided in the reasoning path alone.

\begin{table}[h]
\small
    \centering\scalebox{0.9}{
    \begin{tabular}{p{\linewidth}}
         \toprule
         Let's generate a logically entailed statement from the table \\
         \#\#\# \\
         Table Topic: \textcolor{cyan}{1938 U.S. Open (golf)} \\
         Table Header: \textcolor{brown}{place \# player \# country \# score \# to par \# money} \\
         Table Content: \textcolor{blue}{1 \# ralph guldahl \# united states \# 74 + 70 + 71 + 69 = 284 \# e \# 1000 | ... | 10 \# gene sarazen \# united states \# 74 + 74 + 75 + 73 = 296 \# + 12 \# 106} \\
         Generation: \textcolor{olive}{The majority of the players in the 1938 US Open scored at least 9 over par or above .} \\
         \#\#\# \\
         Table Topic: \textcolor{cyan}{\{table\_topic\}} \\
         Table Header: \textcolor{brown}{\{table\_header\}} \\
         Table Content: \textcolor{blue}{\{table\_content\}} \\
         Generation: \\
         \bottomrule
    \end{tabular}}
    \caption{Example of Direct Prompting for LogicNLG. Each row in the table is separated by a `|' and each entry in a row is separated by a `\#'. The table content is truncated for conciseness.}
    \label{tab:direct_prompt_logicnlg}
\end{table}
\begin{table}[t]
\small
    \centering\scalebox{0.9}{
    \begin{tabular}{p{\linewidth}}
         \toprule
         Let's generate a logically entailed statement from the table for the reasoning path \\
         \#\#\# \\
         Table Topic: \textcolor{cyan}{1938 U.S. Open (golf)} \\
         Table Header: \textcolor{brown}{place \# player \# country \# score \# to par \# money} \\
         Reasoning Path: \textcolor{blue}{most\_greater\_eq \{ all\_rows ; to par ; 9 \}} \\
         Generation: \textcolor{olive}{The majority of the players in the 1938 US Open scored at least 9 over par or above .} \\
         \#\#\# \\
         Table Topic: \textcolor{cyan}{\{table\_topic\}} \\
         Table Header: \textcolor{brown}{\{table\_header\}} \\
         Reasoning Path: \textcolor{blue}{\{reasoning\_path\}} \\
         Generation: \\
         \bottomrule
    \end{tabular}}
    \vspace{-5pt}
    \caption{Example of Surface Realization Prompt for \method{} in LogicNLG.}
    \label{tab:sr_prompt_logicnlg}
\end{table}

\section{Examples of \method{} Summaries}
\label{sec:appendix_examples}
\begin{figure*}[t]
    \centering
    \includegraphics[width=0.6\textwidth]{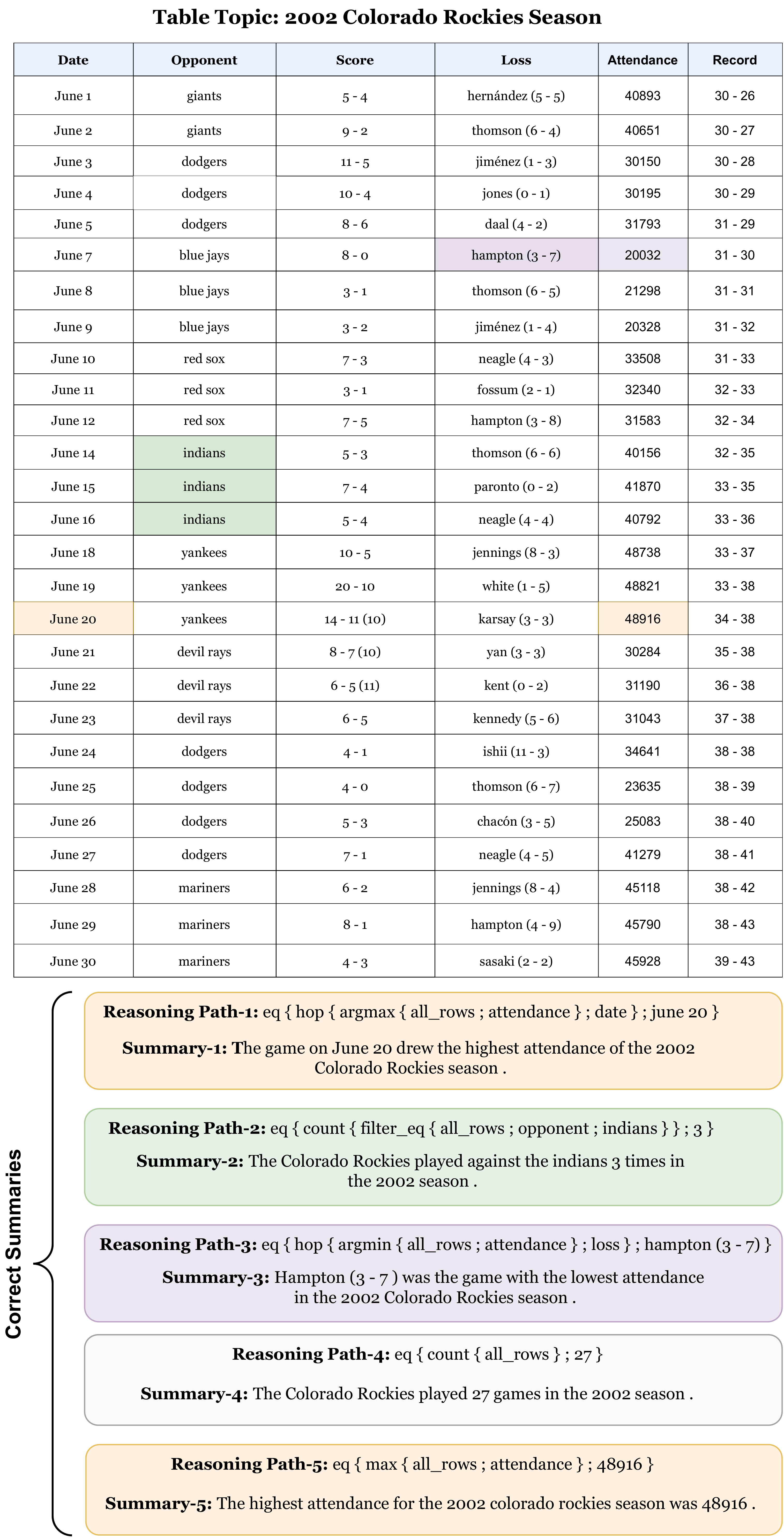}
    \caption{Sample table from LogicNLG dataset and five diverse logical summaries generated by \method{}. Each color code in the table cells highlights parts of the table relevant to a \method{} summary. All generated summaries are logically correct.
    }
    \vspace{-5pt}
    \label{fig:appendix_logicnlg_fig1}
\end{figure*}

\begin{figure*}[t]
    \centering
    \includegraphics[width=0.6\textwidth]{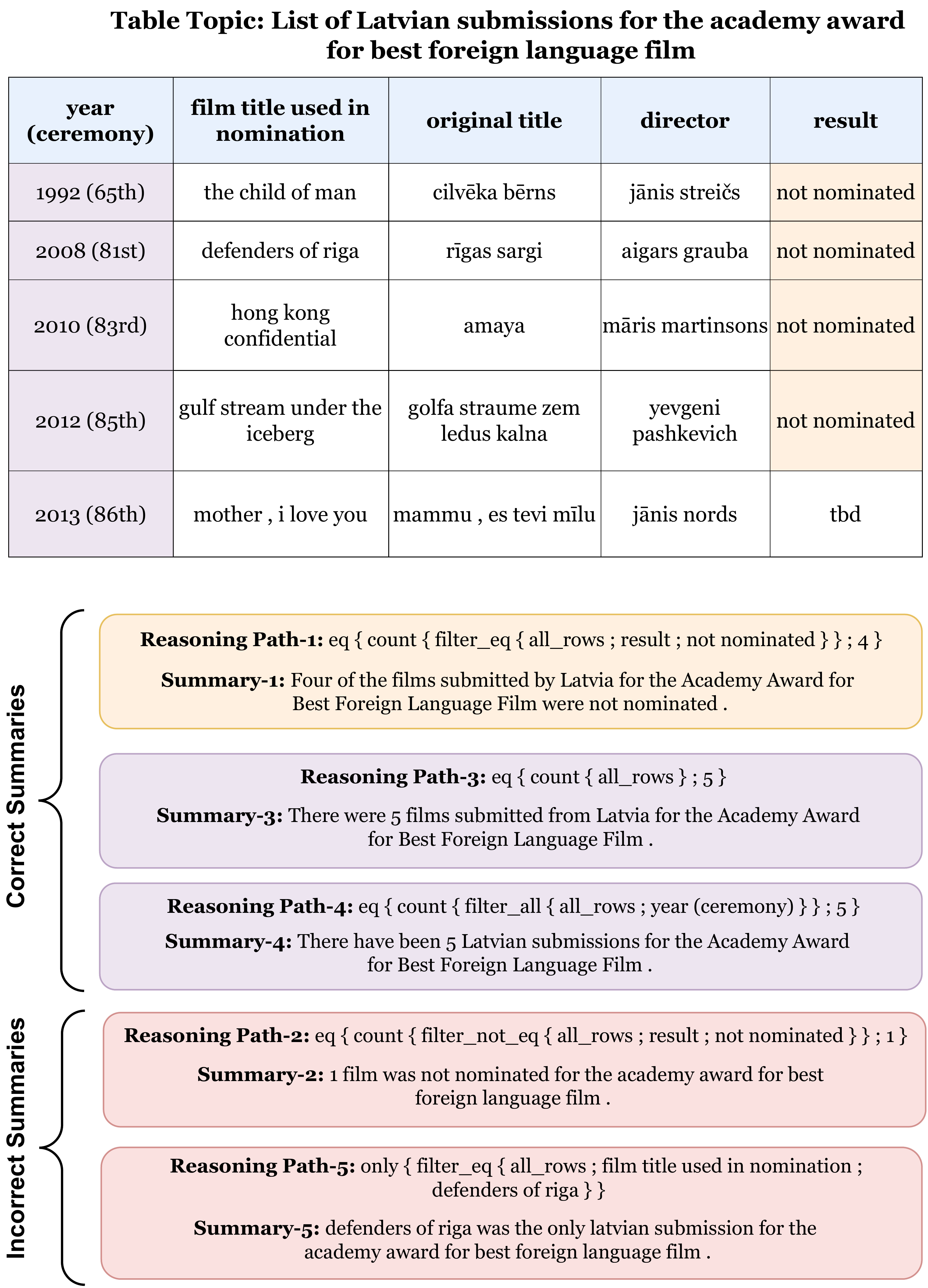}
    \caption{Sample table from LogicNLG dataset and five diverse logical summaries generated by \method{}. Each color code in the table cells highlights parts of the table relevant to a \method{} summary. The red marked blocks are incorrect summaries generated by \method{}.
    }
    \vspace{-5pt}
    \label{fig:appendix_logicnlg_fig2}
\end{figure*}

In Fig.~\ref{fig:appendix_webnlg_examples}, we show representative examples of summaries generated by Direct Prompting and \method{} for WebNLG. Fig.~\ref{fig:appendix_webnlg_generative_process} shows the step-wise summary generation process of \method{} for WebNLG. In Fig.~\ref{fig:appendix_logicnlg_fig1} and \ref{fig:appendix_logicnlg_fig2}, we show representative examples of the reasoning paths and summaries generated for two tables in LogicNLG.

\begin{figure*}[t]
\centering
\small
\resizebox{\textwidth}{!}{%
\begin{tabular}{l p{0.8\linewidth}}
\toprule
\textbf{Input} & \textcolor{olive}{Arlington,\_Texas | isPartOf | Texas} \# Texas | largestCity | Houston \# \textcolor{olive}{Texas | language | English\_language} \\ \midrule
\textbf{Direct Prompting} & Texas is {\textcolor{LightRed}{the second largest state}} in the United States of America. {\textcolor{LightRed}{Its capital}} is Austin and the largest city is Houston. (\textbf{Omissions = 2})  \\ \midrule
\textbf{\method{}} & Houston is the largest city in Texas where English is the official language and Arlington , Texas is a city in Texas. (\textbf{Omissions = 0}) \\
\midrule \midrule
\textbf{Input} & \textcolor{olive}{Hays\_County\_Texas | countySeat | San\_Marcos,\_Texas} \# Austin,\_Texas | isPartOf | Hays\_County,\_Texas \# \textcolor{olive}{Texas | language | Spanish\_language} \# \textcolor{olive}{Austin,\_Texas | isPartOf | Texas} \# \textcolor{olive}{Texas | largestCity | Houston} \\ \midrule
\textbf{Direct Prompting} & Hays County, Texas is a county in the U.S. state of Texas. (\textbf{Omissions = 4}) \\ \midrule
\textbf{\method{}} & Spanish is the official language of Texas. San Marcos is the county seat of Hays County, Texas where Houston is the largest city in Texas {\textcolor{blue}{where}} Austin, Texas is part of Texas. (\textbf{Omissions = 0}) \\ \midrule \midrule
\textbf{Input} & \textcolor{olive}{Vermont | largestCity | Burlington\_Vermont} \# \textcolor{olive}{Alvah\_Sabin | region | Vermont} \# Alvah\_Sabin | activeYearsEndDate | 1857-03-03 \\ \midrule
\textbf{Direct Prompting} & Alvah Sabin \textcolor{red}{was born on March 3, 1857 in Vermont}. (\textbf{Omissions = 2}) \\ \midrule
\textbf{\method{}} & Burlington is the largest city in Vermont where Alvah Sabin is from and he \textcolor{red}{played from 1857-03-03 to 1857-03-03}. (\textbf{Omissions = 0})\\
\bottomrule      
\end{tabular}
}
\caption{Examples of summaries generated by Direct Prompting and \method{} for WebNLG. Hallucinations are marked in \textcolor{red}{red}, omissions are marked in \textcolor{olive}{olive}, and disfluencies are marked in \textcolor{blue}{blue}. Omission count of triples is shown in brackets next to the generations.}
\label{fig:appendix_webnlg_examples}

\end{figure*}

\begin{figure*}[]
\small
\begin{tabular}{l l p{0.38\linewidth} p{0.35\linewidth}}
\toprule
\multicolumn{4}{p{\linewidth}}{\textbf{Input Graph}: Antwerp\_International\_Airport | operatingOrganisation | Flemish\_Government \# Antwerp\_International\_Airport | elevationAboveTheSeaLevel\_(in\_metres) | 12.0 \# Antwerp\_International\_Airport | owner | Flemish\_Region \# Antwerp\_International\_Airport | runwayLength | 600.0} \\ \midrule \midrule
\textbf{\#Step} & \textbf{Module} & \textbf{Input} & \textbf{Output} \\ \midrule
    1 & Surface Realization  &  \textcolor{blue}{Antwerp\_International\_Airport | runwayLength | 600.0}      & \textcolor{brown}{The runway at Antwerp International Airport is 600.0 metres long.}        \\ \midrule
     2  & Surface Realization & \textcolor{blue}{Antwerp\_International\_Airport | elevationAboveTheSeaLevel\_(in\_metres) | 12.0}     &   	\textcolor{brown}{Antwerp International Airport is 12.0 metres above sea level.}     \\\midrule
     3  & Surface Realization & \textcolor{blue}{Antwerp\_International\_Airport | operatingOrganisation | Flemish\_Government}     &   	\textcolor{brown}{Antwerp International Airport is operated by the Flemish Government.}     \\ \midrule
     4 & Text Fusion & \textcolor{blue}{Antwerp International Airport is owned by the Flemish Region.}; \textcolor{olive}{Antwerp International Airport is operated by the Flemish Government.} & \textcolor{brown}{Antwerp International Airport is owned by the Flemish Region and operated by the Flemish Government.} \\ \midrule
     5 & Text Fusion & \textcolor{blue}{Antwerp International Airport is 12.0 metres above sea level.}; \textcolor{olive}{Antwerp International Airport is owned by the Flemish Region and operated by the Flemish Government.} & \textcolor{brown}{Antwerp International Airport is 12.0 metres above sea level and is owned by the Flemish Region and operated by the Flemish Government.} \\ \midrule
     6 & Text Fusion & \textcolor{blue}{The runway at Antwerp International Airport is 600.0 metres long.}; \textcolor{olive}{Antwerp International Airport is 12.0 metres above sea level and is owned by the Flemish Region and operated by the Flemish Government.} & \textcolor{brown}{The runway at Antwerp International Airport is 600.0 metres long and is 12.0 metres above sea level and is owned by the Flemish Region and operated by the Flemish Government.}\\ \midrule
\end{tabular}
\caption{Illustration of the step-wise summary generation process of \method{} for WebNLG. Each step consists of a module (Surface Realization or Fusion), the input to the module (a triple or a pair of texts) and the output summary (Cont. from \cref{sec:value_func}).}
\label{fig:appendix_webnlg_generative_process}
\end{figure*}

\end{document}